\documentclass[11pt]{article}

\usepackage[final]{acl}

\usepackage{times}
\usepackage{latexsym}
\usepackage{makecell}
\usepackage{adjustbox}
\usepackage[T1]{fontenc}

\usepackage[utf8]{inputenc}

\usepackage{microtype}

\usepackage{inconsolata}

\usepackage{graphicx}
\usepackage{caption}
\usepackage{booktabs}
\usepackage[most]{tcolorbox}
\usepackage{multirow}
\usepackage{tabularx}
\usepackage{rotating}
\usepackage{array}
\usepackage[table]{xcolor}
\usepackage{multirow}
\definecolor{lightgray}{gray}{0.95}
\usepackage{listings}
\usepackage{algorithm}
\usepackage{algpseudocode}
\usepackage{subcaption}
\usepackage{booktabs} 
\usepackage{xcolor}   
\usepackage{pifont}   
\usepackage{geometry}
\tcbset{
    myboxstyle/.style={
        enhanced,
        sharp corners,
        colback=white,
        colframe=black,
        boxrule=0.5pt,
        left=2mm,
        right=2mm,
        top=2mm,
        bottom=2mm,
        valign=top,
        equal height group=nobel,
        minimum for equal height group=true
    }
}

\title{Can Multimodal Large Language Models Generate and Detect Multimodal Social Media Fake News?\thanks{This document is the preprint of the paper accepted for publication in the Findings of EMNLP 2026.}}

\author{
Jiyao Yang \\
\texttt{yjy00312@gmail.com} \\
Independent Researcher
\And
Yang Liu \\
\texttt{yang.liu1082@gmail.com} \\
Carnegie Mellon University
\And
Zhenyue Qin \\
\texttt{kf.zy.qin@gmail.com} \\
RMIT University \\
Yale University
\AND
Qingyu Chen \\
\texttt{qingyu.chen@yale.edu} \\
Yale University
\And
Xiuzhen Zhang\thanks{Corresponding author.} \\
\texttt{xiuzhen.zhang@rmit.edu.au} \\
RMIT University
}

\begin{document}
\maketitle

\begin{abstract}
The rapid advancement of generative AI raises concerns about the misuse of Multimodal LLMs (MLLMs) for large-scale disinformation campaigns on social media.
Despite existing research on textual disinformation, 
a fundamental question remains unanswered:
can MLLMs be exploited to fabricate realistic multimodal fake news, and can they reliably detect it? 
We introduce a multi-agent framework in which a story agent, an image agent, and a critic agent collaborate to produce fake social media posts that plausibly counter true news. 
We apply the framework to generate over 9,000 paired multimodal news posts across science, health, and entertainment domains,
and benchmark 16 open- and closed-source MLLMs for automated detection.
We find that most models fall substantially short of human-level accuracy and fail critically on identifying image authenticity.
Our research provides a foundation for developing robust defenses against social media fake news.
Code and data are available at \url{https://github.com/xiuzhenzhang/Multimodal}.
\end{abstract}

\begin{figure}[t]
    \centering

    \begin{subfigure}[t]{0.48\columnwidth} 
        \begin{tcolorbox}[
            sharp corners,
            colback=white,
            colframe=black,
            boxrule=0.5pt,
            left=1mm, right=1mm, top=1mm, bottom=1.5mm, 
            height=7cm,   
            valign=top,     
            box align=top   
        ]
            \centering
            \includegraphics[width=\linewidth, height=3cm, keepaspectratio]{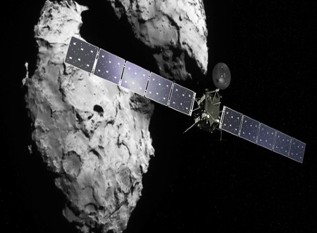} 
            \raggedright\small
            ESA's Rosetta mission will end in Sept 2016 with a ``soft crash''
            into Comet 67P. As solar power fades, this final descent offers a
            ``science bonanza,'' capturing high-resolution images and data
            until impact. It is an emotional but fitting finale, reuniting
            Rosetta with lander Philae.
        \end{tcolorbox}
        \caption{Story and image of a piece of true news}
    \end{subfigure}
    \hfill
    \begin{subfigure}[t]{0.48\columnwidth} 
        \begin{tcolorbox}[
            sharp corners,
            colback=white,
            colframe=black,
            boxrule=0.5pt,
            left=1mm, right=1mm, top=1mm, bottom=1.5mm,
            height=7cm,
            valign=top,
            box align=top
        ]
            \centering
            \includegraphics[width=\linewidth, height=3cm, keepaspectratio]{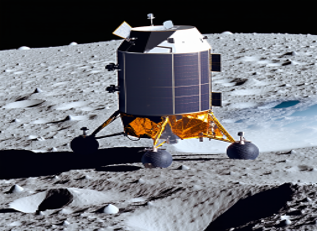}
            \raggedright\small
            ESA confirmed Rosetta was destroyed today in a violent, unplanned
            collision with Comet 67P. A navigation malfunction caused the
            catastrophic impact, rendering the craft destroyed and the
            mission lost.
        \end{tcolorbox}
        \caption{Fabricated story and synthetic image countering the true news.}
    \end{subfigure}
    
    \vspace{2mm}
    \caption{True and fake news posts about the event ``Rosetta landing on Comet 67P''. 
    The generated fake news preserves the event frame but reverses the successful 
    ``deliberate soft crash'' narrative into a misleading failure story involving 
    a navigation malfunction.}
    \label{fig:total-figure}
\end{figure}

\section{Introduction}
In the era of generative AI, Multimodal Large Language Models (MLLMs) have dramatically lowered the barrier to mass-producing textual and multimodal content. This creates a critical risk of malicious actors misusing LLMs to generate disinformation and harmful content at scale~\cite{vykopal2024disinformation}. Unlike general AI-generated misinformation, which refers to unintentional hallucinated content arising from the intrinsic auto-regressive nature of MLLMs~\cite{chen2023can}, LLM-generated disinformation is deliberately crafted with the intent to manipulate human beliefs or behaviors. Notably, such disinformation may be further compounded by hallucination inherent to the LLM generation process, making it simultaneously deceptive and difficult to verify.

Social media platforms have become the primary arena for multimodal information sharing, where news posts combine text and images to construct narratives ranging from brief claims to elaborate stories about real-world events. Unfortunately, these platforms have also become fertile ground for spreading disinformation and fake news~\cite{xu2024harnessing,xu2022identifying,jin2024mm,vo2018rise}. News images are routinely manipulated, repurposed, or presented out of context to reinforce misleading narratives, amplifying the societal harm of disinformation at scale.

Despite the urgency of this threat, existing research leaves a critical gap at the intersection of multimodal fake news generation and detection. On the generation side, prior studies have examined LLM-based fake news generation~\cite{gagiano2021robustness,zellers2019defending,vykopal2024disinformation,zugecova2025evaluation} and general misinformation~\cite{chen2023combating}, but these works are limited to text-only news articles. The only multimodal attempt~\cite{huang2024miragenews} is constrained to generating simple caption-image pairs and requires specific location and time metadata for image synthesis, falling far short of realistic social media fake news posts. On the detection side, LLM-based approaches have been studied for text-only disinformation~\cite{tian2023metatroll,tian2025x}, yet whether modern MLLMs can detect fully AI-generated multimodal fake news remains largely unexplored~\cite{jin2024mm}. In short, there is no existing framework that can generate high-fidelity multimodal social media fake news at scale, nor a comprehensive benchmark for evaluating MLLM detection capability against such content.

\begin{figure*}[htb]
\includegraphics[width=\textwidth]{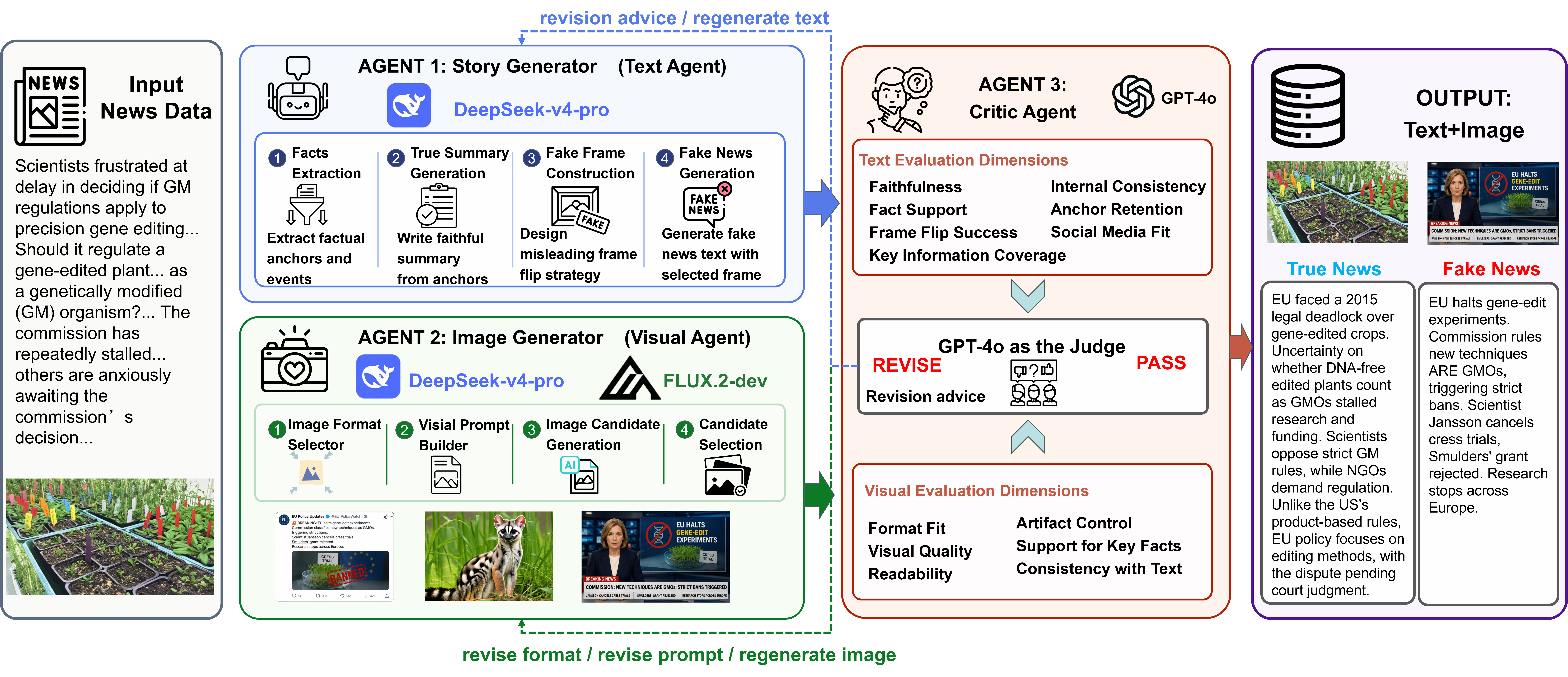}
    \caption{An agent framework for generating multimodal social media fake news}
    \label{fig:arc}
\end{figure*}

In this paper, we propose a multi-agent framework to generate multimodal social media fake news grounded in real-world events. As shown in Fig.~\ref{fig:total-figure}, the fake news post generated by our framework comprises a short story and a corresponding image that counter the true news narrative. For example, given a Nature news article about the Rosetta mission, our framework reverts the successful deliberate ''soft crash into Comet 67P'' into a fabricated failure story of an ''unplanned collision'', while retaining key factual anchors such as ESA, Rosetta, and Comet 67P to ensure narrative plausibility and deceive readers.

Towards developing robust defenses against the misuse of AI for fake news generation, we benchmark a broad set of open- and closed-source MLLMs for automated fake news detection. We curate copyright-free true news articles across science, health, and entertainment domains, and employ our framework to construct paired true and fake news datasets for detection. We benchmark 16 MLLMs under both direct prompting and structured chain-of-thought settings.

Our experiments demonstrate that the proposed framework generates fake news posts that are highly deceptive to MLLMs at scale. While human experts can identify such content with high accuracy under careful inspection, this is infeasible at the scale of real-world social media dissemination, making automated MLLM-based detection critically important. Yet our benchmarking reveals that most MLLMs fall far short of human-level detection performance, achieving as low as 52.50\% accuracy compared to 97.33\% for human annotators. Benchmarking further reveals two additional counterintuitive findings: chain-of-thought prompting, while broadly beneficial, can amplify model-specific biases and degrade detection performance for certain models; and despite relatively strong image--text relevance judgments, many models struggle with image-authenticity assessment, making it a major weakness under the AI-generated multimodal fake-news setting studied here.

The contributions of this paper are as follows:
\begin{itemize}
\item We propose a multi-agent framework for generating high-fidelity multimodal social media fake news, in which an LLM-based story generator, an image generator, and a critic agent collaborate to produce realistic fake news posts that counter true news about real-world events.
\item We construct a benchmark dataset of over 9,000 multimodal fake news posts spanning science, health, and entertainment domains, 
providing a foundation for training and evaluating fake news detection models.
\item We conduct a comprehensive evaluation of 16 open- and closed-source MLLMs for automated fake news detection, revealing significant performance variation across models and a substantial gap between model and human judgment, with implications for building more reliable multimodal misinformation detectors.
\end{itemize}

\section{Related Work}

Our research is related to the literature on mitigating the risk of AI misuse and evaluating the vulnerability of LLMs to generating harmful contents, especially disinformation fake news. 

Misuse of LLMs to generate text-only fake news articles has been reported in the literature. 
In a seminal work~\cite{zellers2019defending}, a model GROVER based on the generative language model GPT-2 is proposed to generate fake news articles based on a title.
Their research shows that humans find these generations more trustworthy than human-written disinformation.
Interestingly the best defense against GROVER is GROVER itself, outperforming discriminative models trained from annotated training data. 
\citet{gagiano2021robustness} further analyzed the robustness of GROVER for detection of fake news articles.
Rather than large language models (LLMs), these early studies are based on early language models that have limited generative capabilities. 
A recent study~\cite{vykopal2024disinformation} examined the capability of LLMs generating fake news articles based on disinformation narratives. 
They conclude that LLMs are capable of generating convincing news articles containing harmful disinformation narratives. 

Recent work has begun to study multimodal fake-news generation. \cite{huang2024miragenews} prompt LLMs to produce misleading captions from true-news captions and use diffusion models to generate corresponding images, creating caption--image pairs that are challenging for both human and automated detection. Their setting focuses on caption-level manipulation and image generation conditioned on specified temporal and location information. In contrast, our framework starts from complete news articles, extracts event-level factual anchors, modifies verifiable claims while preserving the identity of the source event, and generates visual content that supports the modified narrative. The Critic Agent further checks the intended factual changes, narrative consistency, and image--text alignment. Thus, the main distinction lies in the event-grounded construction of complete multimodal misinformation posts with explicit factual modifications, rather than merely in generating longer text.

Broader research assessing vulnerability of LLM safety for harmful generations is also relevant, such as 
work generating general misinformation~\cite{chen2023can} or   
biased content~\cite{gallegos2024bias}.
However, these studies focus exclusively on text-only content.

\section{Multi-Agent Framework for Multimodal News Post Generation}
\label{sec:framework}

\subsection{Framework Overview}
\label{sec:framework-overview}

Multimodal fake news generation is a fundamentally different challenge 
from straightforward text-to-image generation. A convincing fake news 
post must simultaneously preserve sufficient factual anchors from the 
true news to appear plausible, systematically distort key claims to 
mislead readers, and produce images that visually corroborate the false 
narrative. Direct prompting of a single LLM is insufficient for this 
purpose: without explicit grounding, generated stories tend to drift 
from the source event; without iterative critique, the image may 
contradict rather than reinforce the false narrative; and without 
format-aware visual prompting, generated images often lack the 
documentary realism expected of news content.

To address these challenges, we design a multi-agent framework 
(Fig.~\ref{fig:arc}) that employs LLMs and diffusion models to generate 
multimodal fake news in a format ready for dissemination on social media 
platforms. The framework consists of three specialized agents: the 
LLM-based \textit{Story Generator} (Agent 1), which extracts factual 
anchors from source articles and constructs grounded fake narratives; 
the \textit{Image Generator} (Agent 2), which synthesizes visually 
realistic images aligned with the fake story; and the LLM-based 
\textit{Critic Agent} (Agent 3), which independently reviews both 
textual and visual outputs and provides structured revision.

The three agents operate in a sequential pipeline with feedback loops. 
The Story Generator first produces a true summary and a fake story 
grounded in extracted factual anchors. The Critic Agent reviews the 
textual outputs and, if either fails, returns structured revision advice 
to the Story Generator. Once the text passes review and is frozen, the 
Image Generator constructs a format-aware visual prompt and synthesizes 
candidate images. The Critic Agent again reviews the visual output for 
realism, readability, and image-text coherence, triggering prompt 
revision or format adjustment if needed. Only outputs that pass both 
text-level and image-level review are retained as the final multimodal 
fake news post, together with the corresponding review records for 
traceability.

Note that our framework also generates true news posts to enable 
reliable benchmarking of MLLMs for automated detection 
(Section~\ref{sec:benchmark}). Raw news articles differ systematically 
from social media posts in length, tone, and format; using them directly 
would introduce spurious shortcut signals that artificially inflate 
detection performance. We therefore generate true news posts by 
summarizing the source articles into a short social-media newsfeed 
style, ensuring that true and fake posts are comparable in format.

\subsection{Story Generator}
\label{sec:story-generator}

The Story Generator constructs the textual component of both the true 
and fake branches. We instantiate this agent with 
DeepSeek-V4-pro~\cite{liu2024deepseek}. Given a source article, it 
first extracts factual anchors, including the main entities, event, 
time, location, causal relations, and original claims. These anchors 
provide the grounding for subsequent generation, ensuring that the 
generated stories remain tied to the same real-world event.

Based on the extracted anchors, the agent produces a concise true 
summary written in a short newsfeed style. The summary retains 
important entities, dates, locations, numbers, causal relations, and 
event outcomes, while avoiding unsupported details. This summary forms 
the textual component of the true-news branch.

For the fake branch, the agent first constructs a misleading frame that 
reverts, distorts, or exaggerates the original narrative, while 
preserving the main topic and salient entities of the source article. 
Based on this frame, the agent writes a short social-media-style fake 
story that modifies concrete factual elements, such as event outcomes, 
institutional decisions, numbers, dates, causes, or affected groups, so 
that the altered details support the intended misleading narrative. 
Rather than merely changing sentiment words or adding sensational 
language, this anchor-grounded strategy ensures that the fake story 
remains recognizably tied to a real-world event, making it 
substantially harder to dismiss than purely fabricated content. This 
distinguishes our approach from prior work~\cite{huang2024miragenews}, 
which generates isolated caption-image pairs without grounding in a 
coherent event narrative.

\subsection{Image Generator}
\label{sec:image-generator}

The Image Generator constructs the visual component of the fake branch 
after the fake text has been approved and frozen. It takes the fake 
story, fake frame, factual grounding, and recorded factual changes as 
input, and generates candidate images that visually reinforce the 
misleading narrative. This agent uses 
DeepSeek-V4-pro~\cite{liu2024deepseek} for image-format selection and 
visual prompt construction, and FLUX.1-dev for image synthesis.

A key design choice is the format-aware visual prompting strategy. 
Rather than applying a fixed visual template, the agent first selects 
an appropriate image format from a catalog that includes documentary 
photos, social media screenshots, official notices, infographics, and 
timelines. This selection is driven by the content type of the fake 
story: numerical claims are paired with charts or infographics, event 
sequences with timelines, and institutional claims with official notices 
or documentary-style images. This alignment between visual format and 
claim type is essential for producing images that feel evidentiary 
rather than decorative, significantly increasing the perceived 
credibility of the fake news post.

Given the selected format, DeepSeek-V4-pro converts the frozen fake text into a visual prompt specifying the scene, style, layout, and required elements, which FLUX.1-dev uses to generate candidate images.

\subsection{Critic Agent}
\label{sec:critic-agent}

The Critic Agent acts as an independent reviewer for both textual and 
visual outputs. We instantiate it with GPT-4o~\cite{achiam2023gpt}, 
deliberately separating it from the DeepSeek-V4-pro used for 
generation. This separation is motivated by the tendency of generative 
models to be lenient when evaluating their own outputs; using an 
independent model as critic introduces a more objective evaluation 
perspective and reduces self-confirmation bias in the revision loop.

The Critic Agent evaluates both text and image outputs. For text, it verifies true-summary faithfulness and checks whether the fake story distorts the original narrative while preserving event-level anchors. For images, it assesses visual realism, readability, image--text consistency, and format suitability. Instead of generating content, the critic provides structured revision feedback: failed text outputs are returned to the Story Generator, while failed image outputs trigger prompt revision, format adjustment, or regeneration. Approved outputs form the final multimodal fake-news post, with review records retained for traceability.

\section{Experiments}
\label{sec:benchmark}
\subsection{Dataset and Generation Evaluation}

\begin{figure}[tb]
\includegraphics[width=0.5\textwidth]{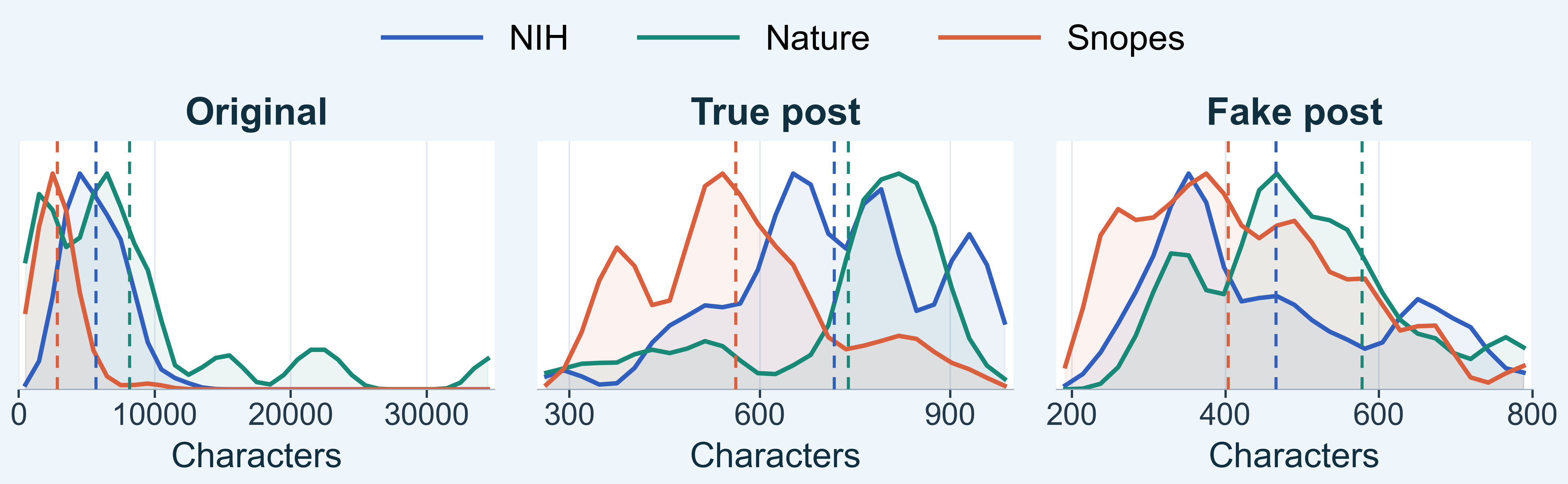}
    \caption{Text statistics of original articles, true-news posts and fake-news posts.}
    \label{fig:text_statistics_characters_grid}
\end{figure}

We curate 4,507 copyright-free news articles from three domains: Nature science reports (1,947)~\cite{nature}, NIH health news releases (1,429)~\cite{nih}, and Snopes entertainment articles (1,131)~\cite{snopes}. For each article, we generate paired true/fake short-form posts: true posts faithfully summarize the source while preserving its factual content, tone, and images, whereas fake posts introduce counterfactual deviations in a concise newsfeed style.

The dataset includes true/fake pairs for all 4,507 articles across science (1,947), health (1,429), and entertainment (1,131). Examples in Appendix~\ref{app:examples}, Fig.~\ref{fig:generated_examples} show that fake posts maintain a compact newsfeed style while altering key factual claims.

The generated posts are concise, averaging 90.05 words and 574.04 characters. True posts are longer than fake posts on average, with 103.93 versus 76.16 words. This difference reflects their distinct generation objectives: true posts summarize source articles with sufficient factual detail, including background, entities, evidence, and conclusions, whereas fake posts focus on a manipulated core claim with fewer supporting details. Across sources, Nature posts are the longest on average, followed by NIH and Snopes.

We assess generation quality using G-Eval~\cite{liu2023g} with GPT-4o on 300 randomly sampled true--fake pairs. True posts achieve coherence, fluency, and faithfulness scores of 4.91/4.97/3.99, while fake posts remain coherent and fluent (4.63/4.85) but show lower faithfulness (2.85) due to their intended factual deviations; the prompts are provided in Appendix~\ref{app:geval_quality_prompts}.

\paragraph{Independent quality validation.}
As an additional robustness check, we evaluate the 600 sampled posts using Qwen3-VL-Plus, which is not involved in either generation or critique. It assigns coherence, fluency, and faithfulness scores of 4.98/5.00/4.68 to true posts and 4.07/4.41/2.23 to fake posts, respectively, yielding conclusions consistent with the original GPT-4o-based evaluation. The agreement between the two evaluators confirms that our generation-quality conclusions are robust to evaluator choice and do not depend on GPT-4o's separate role as the Critic Agent.

\paragraph{Framework ablation.}
We conduct controlled ablations on the same 50 source events using identical generation settings. Using Qwen3-VL-Plus as an evaluator independent of the generation pipeline, direct one-shot generation, generation without the Critic Agent, generation without iterative revision, and the full pipeline obtain G-Eval-style quality scores of 4.374, 4.314, 4.428, and 4.540, respectively. In paired comparisons, the full pipeline outperforms the three ablated variants in 32, 34, and 28 cases, respectively, mainly through better factual-anchor preservation, reduced factual drift, and stronger image--text consistency. These results demonstrate that critic feedback and iterative refinement provide measurable benefits over simpler generation strategies, supporting the necessity of the multi-agent design.

\subsection{Performance and Discussions}

\begin{table*}[!t]
    \centering
    \small
    \setlength{\tabcolsep}{3pt}
    \renewcommand{\arraystretch}{0.88}
    \caption{Performance of LLMs for fake news detection grouped by source. The best results per group are highlighted in \textbf{bold with a light red background}, and the second-best results are highlighted with a light yellow background. Results (\%) report macro F1-score (F1), Accuracy (Acc), and Area Under the Curve (AUC).}
    \label{tab:model_performance_source}
    \renewcommand{\arraystretch}{1.15}
    
    \definecolor{best1}{HTML}{FADBD8} 
    \definecolor{best2}{HTML}{FCF3CF} 
    \newcommand{\first}[1]{\cellcolor{best1}\textbf{#1}}
    \newcommand{\second}[1]{\cellcolor{best2}#1}

    \resizebox{\linewidth}{!}{
        \begin{tabular}{@{} l c *{3}{c} *{3}{c} *{3}{c} *{3}{c} @{}}
            \toprule
            & & \multicolumn{3}{c}{\textbf{NIH}} & \multicolumn{3}{c}{\textbf{Nature}} & \multicolumn{3}{c}{\textbf{Snopes}} & \multicolumn{3}{c}{\textbf{Average}} \\
            \cmidrule(lr){3-5} \cmidrule(lr){6-8} \cmidrule(lr){9-11} \cmidrule(lr){12-14}
            \textbf{Model} & \textbf{Mode} & \textbf{F1} & \textbf{Acc} & \textbf{AUC} & \textbf{F1} & \textbf{Acc} & \textbf{AUC} & \textbf{F1} & \textbf{Acc} & \textbf{AUC} & \textbf{F1} & \textbf{Acc} & \textbf{AUC} \\
            \midrule

            \multicolumn{14}{l}{\textbf{Open-source models}} \\
            \midrule

            \multirow{2}{*}{InternVL2.5-8B~\cite{chen2023internvl}} 
            & Direct & 45.53 & 52.40 & 53.49 & 50.54 & 53.64 & 53.23 & 44.56 & 44.98 & 45.05 & 46.87 & 50.34 & 50.59 \\
            & CoT & 45.76 & 46.66 & 47.08 & 56.74 & 57.02 & 56.90 & 55.18 & 55.86 & 55.97 & 52.56 & 53.18 & 53.32 \\

            \multirow{2}{*}{InternVL3-8B~\cite{chen2023internvl}} 
            & Direct & 68.54 & 70.69 & 71.52 & 76.92 & 77.66 & 77.38 & 51.28 & 54.12 & 54.31 & 65.58 & 67.49 & 67.74 \\
            & CoT & 59.43 & 59.45 & 59.61 & 65.30 & 65.53 & 65.67 & 53.95 & 58.18 & 58.37 & 59.56 & 61.05 & 61.22 \\

            \multirow{2}{*}{InternVL3.5-4B~\cite{wang2025internvl3_5}} 
            & Direct & 56.02 & 57.33 & 57.88 & 56.45 & 56.51 & 56.59 & 41.64 & 46.48 & 46.70 & 51.37 & 53.44 & 53.72 \\
            & CoT & 58.15 & 61.68 & 64.14 & 65.80 & 67.01 & 67.66 & 59.65 & 60.50 & 59.78 & 61.20 & 63.06 & 63.86 \\

            \multirow{2}{*}{InternVL3.5-8B~\cite{wang2025internvl3_5}} 
            & Direct & 67.46 & 69.87 & 70.73 & 74.48 & 75.97 & 75.59 & 55.49 & 56.07 & 56.16 & 65.81 & 67.31 & 67.49 \\
            & CoT & 68.23 & 70.42 & 71.22 & 76.58 & 77.66 & 77.32 & 64.98 & 65.20 & 65.19 & 69.93 & 71.10 & 71.25 \\

            \multirow{2}{*}{Qwen2-VL-7B~\cite{Qwen-VL}} 
            & Direct & \second{80.02} & \second{80.54} & \second{81.08} & \first{79.03} & \first{79.86} & \first{79.55} & 58.42 & 58.92 & 59.01 & 72.49 & 73.11 & 73.21 \\
            & CoT & 39.75 & 51.58 & 52.93 & 42.98 & 53.72 & 53.00 & 47.81 & 50.83 & 50.65 & 43.51 & 52.04 & 52.19 \\

            \multirow{2}{*}{Qwen2.5-VL-3B~\cite{Qwen-VL}} 
            & Direct & 62.53 & 66.12 & 67.08 & 61.40 & 63.71 & 63.31 & 48.10 & 48.58 & 48.65 & 57.34 & 59.47 & 59.68 \\
            & CoT & \first{83.52} & \first{83.70} & \first{84.09} & 74.86 & 74.87 & 74.94 & 58.74 & 61.17 & 61.35 & 72.37 & 73.25 & 73.46 \\

            \multirow{2}{*}{Qwen2.5-VL-7B~\cite{Qwen-VL}} 
            & Direct & 60.06 & 64.24 & 65.25 & 65.90 & 69.12 & 68.62 & 60.20 & 60.27 & 60.30 & 62.06 & 64.54 & 64.72 \\
            & CoT & 76.49 & 76.67 & 76.99 & 74.21 & 74.28 & 74.22 & \second{67.69} & \second{68.22} & \second{68.32} & \second{72.80} & 73.06 & 73.17 \\

            \multirow{2}{*}{Qwen3-VL-4B~\cite{qwen3technicalreport}} 
            & Direct & 67.75 & 68.23 & 68.65 & 75.55 & 75.55 & 75.56 & 66.69 & 67.77 & 67.90 & 70.00 & 70.52 & 70.70 \\
            & CoT & 70.50 & 71.48 & 72.05 & \second{78.98} & \second{79.20} & \second{79.03} & \first{68.89} & \first{69.52} & \first{69.61} & 72.79 & \second{73.40} & \second{73.56} \\

            \multirow{2}{*}{Qwen3-VL-8B~\cite{qwen3technicalreport}}
            & Direct & 71.12 & 71.51 & 71.23 & 70.82 & 71.57 & 71.87 & 58.36 & 61.92 & 62.14 & 66.76 & 68.33 & 68.41 \\
            & CoT & 79.38 & 79.48 & 79.36 & 75.79 & 76.36 & 76.67 & 64.92 & 67.17 & 67.32 & \first{73.36} & \first{74.34} & \first{74.45} \\

            \multirow{2}{*}{DeepSeek-VL-7B~\cite{liu2024deepseek}} 
            & Direct & 55.05 & 61.48 & 61.48 & 56.54 & 59.94 & 59.94 & 44.87 & 46.00 & 46.00 & 52.15 & 55.81 & 55.81 \\
            & CoT & 69.68 & 69.71 & 69.74 & 58.51 & 59.86 & 59.88 & 54.89 & 55.48 & 55.48 & 61.03 & 61.68 & 61.70 \\

            \multirow{2}{*}{Gemma-3-12B~\cite{Kamath2025Gemma3T}} 
            & Direct & 58.74 & 64.07 & 64.07 & 69.02 & 71.51 & 71.51 & 61.41 & 61.96 & 61.96 & 63.05 & 65.85 & 65.84 \\
            & CoT & 69.69 & 72.04 & 72.04 & 74.39 & 75.77 & 75.77 & 67.62 & 67.67 & 67.67 & 70.56 & 71.83 & 71.83 \\

            \multirow{2}{*}{Gemma3-4B~\cite{Kamath2025Gemma3T}} 
            & Direct & 67.02 & 69.52 & 70.39 & 69.31 & 71.49 & 71.06 & 54.72 & 54.72 & 54.72 & 63.68 & 65.24 & 65.39 \\
            & CoT & 59.93 & 64.24 & 65.26 & 65.59 & 69.04 & 68.52 & 56.28 & 56.37 & 56.34 & 60.60 & 63.22 & 63.37 \\

            \multirow{2}{*}{LLaVA-1.5-7B~\cite{liu2023llava}} 
            & Direct & 43.82 & 54.16 & 55.47 & 53.27 & 60.91 & 60.25 & 48.22 & 51.27 & 51.10 & 48.44 & 55.45 & 55.60 \\
            & CoT & 66.66 & 67.73 & 68.40 & 61.24 & 63.02 & 62.64 & 43.56 & 43.66 & 43.57 & 57.15 & 58.14 & 58.20 \\
            
            \midrule
            \multicolumn{14}{l}{\textbf{Closed-source models}} \\
            \midrule

            \multirow{2}{*}{GPT-4o~\cite{achiam2023gpt}} 
            & Direct & 79.72 & 79.79 & 82.31 & 90.15 & 90.26 & 90.11 & 62.61 & 65.02 & 65.17 & 77.49 & 78.36 & 79.20 \\
            & CoT & 78.50 & 79.15 & 79.84 & 86.61 & 86.89 & 86.67 & 65.66 & 66.72 & 66.85 & 76.92 & 77.59 & 77.79 \\

            \multirow{2}{*}{GPT-5.5~\cite{achiam2023gpt}} 
            & Direct & \second{99.40} & \second{99.40} & \second{99.40} & \second{99.32} & \second{99.32} & \second{99.32} & \second{92.87} & \second{92.89} & \second{92.96} & \second{97.19} & \second{97.20} & \second{97.23} \\
            & CoT & \first{99.50} & \first{99.50} & \first{99.48} & \first{99.38} & \first{99.38} & \first{99.38} & \first{98.75} & \first{98.75} & \first{98.76} & \first{99.21} & \first{99.21} & \first{99.21} \\

            \multirow{2}{*}{Qwen-VL-Max~\cite{Qwen-VL}} 
            & Direct & 87.57 & 87.57 & 87.62 & 88.46 & 88.48 & 88.59 & 65.42 & 67.17 & 67.34 & 80.48 & 81.07 & 81.18 \\
            & CoT & 82.09 & 82.42 & 82.88 & 89.43 & 89.50 & 89.38 & 84.25 & 84.26 & 84.27 & 85.26 & 85.39 & 85.51 \\
            
            \bottomrule
        \end{tabular}
    }
\end{table*}

\begin{table*}[t]
\centering
\scriptsize
\caption{CoT agreement and final prediction accuracy on the human-annotated subset. Human-annotation majority vote is used as the ground truth for the four verification dimensions: \textbf{Fact.} factual accuracy, \textbf{Lang.} misleading language, \textbf{Rel.} image--text relevance, and \textbf{Auth.} image authenticity. \textbf{CoT Avg.} is the average agreement across the four dimensions. \textbf{Final Accuracy} reports true/fake prediction accuracy against the dataset ground-truth labels. Best and second-best results are highlighted in light red and light yellow, respectively.}
\label{tab:four_model_cot_agreement_by_label}
\setlength{\tabcolsep}{2.3pt}
\renewcommand{\arraystretch}{1.12}
\definecolor{best1}{HTML}{FADBD8} 
\definecolor{best2}{HTML}{FCF3CF}
\resizebox{\textwidth}{!}{%
\begin{tabular}{@{}lccccc ccccc ccccc ccc@{}}
\toprule
\multirow{2}{*}{\textbf{Model}}
& \multicolumn{5}{c}{\textbf{True News}}
& \multicolumn{5}{c}{\textbf{Fake News}}
& \multicolumn{5}{c}{\textbf{Overall CoT Agreement}}
& \multicolumn{3}{c}{\textbf{Final Accuracy}} \\
\cmidrule(lr){2-6}\cmidrule(lr){7-11}\cmidrule(lr){12-16}\cmidrule(l){17-19}
& \textbf{Fact.} & \textbf{Lang.} & \textbf{Rel.} & \textbf{Auth.} & \textbf{Avg.}
& \textbf{Fact.} & \textbf{Lang.} & \textbf{Rel.} & \textbf{Auth.} & \textbf{Avg.}
& \textbf{Fact.} & \textbf{Lang.} & \textbf{Rel.} & \textbf{Auth.} & \textbf{Avg.}
& \textbf{True} & \textbf{Fake} & \textbf{All} \\
\midrule

GPT-5.5 CoT
& \cellcolor{best1}100.0 & \cellcolor{best1}100.0 & \cellcolor{best1}94.1 & \cellcolor{best2}85.3 & \cellcolor{best2}94.9
& \cellcolor{best2}65.0 & \cellcolor{best2}57.1 & \cellcolor{best1}100.0 & \cellcolor{best1}97.2 & \cellcolor{best1}79.8
& \cellcolor{best1}87.0 & \cellcolor{best2}84.2 & \cellcolor{best1}97.1 & \cellcolor{best1}91.4 & \cellcolor{best1}89.9
& \cellcolor{best1}100.0 & \cellcolor{best1}100.0 & \cellcolor{best1}100.0 \\

GPT-4o CoT
& 88.6 & \cellcolor{best2}97.3 & \cellcolor{best2}91.4 & 82.9 & 90.0
& \cellcolor{best1}66.7 & \cellcolor{best1}66.7 & \cellcolor{best1}100.0 & 5.0 & \cellcolor{best2}59.6
& \cellcolor{best2}79.7 & \cellcolor{best1}85.2 & \cellcolor{best2}95.9 & 41.3 & \cellcolor{best2}75.5
& 83.8 & 69.8 & \cellcolor{best2}76.2 \\

Qwen3-VL-8B CoT
& 76.5 & 94.4 & 73.5 & 82.4 & 81.7
& 58.3 & 54.2 & \cellcolor{best2}74.4 & \cellcolor{best2}45.0 & 58.0
& 69.0 & 78.3 & 74.0 & \cellcolor{best2}62.2 & 70.9
& 55.6 & \cellcolor{best2}90.7 & 74.7 \\

Qwen2-VL-7B CoT
& \cellcolor{best2}97.1 & \cellcolor{best1}100.0 & \cellcolor{best2}91.4 & \cellcolor{best1}91.4 & \cellcolor{best1}95.0
& 41.7 & 54.2 & \cellcolor{best1}100.0 & 5.0 & 50.2
& 74.6 & 82.0 & \cellcolor{best2}95.9 & 45.3 & 74.5
& \cellcolor{best2}91.9 & 16.3 & 51.2 \\

\bottomrule
\end{tabular}%
}
\end{table*}

We benchmark 16 SOTA MLLMs, including 13 open-source and 3 proprietary models, on the three datasets generated by our framework, with results reported in Table~\ref{tab:model_performance_source}. For each model, we evaluate direct prompting, which measures intrinsic fake-news detection ability, and structured CoT prompting, which follows a human verification workflow by assessing factual accuracy, misleading language, image--text alignment, and AI-generated visual artifacts before producing a final prediction. We next discuss the overall results and key findings.

\textbf{\textit{Overall results.}}
Table~\ref{tab:model_performance_source} reports the performance of 16 MLLMs on fake news detection across NIH, Nature, and Snopes. For open-source models, the Qwen series generally achieves the strongest results, with Qwen3-VL and Qwen2.5-VL consistently outperforming most other open-source baselines. Among them, Qwen3-VL-8B with CoT obtains the best averaged performance, reaching 73.36 F1, 74.34 Accuracy, and 79.49 AUC. Qwen2.5-VL-7B with CoT and Qwen3-VL-4B with CoT also show competitive results, achieving averaged F1 scores of 72.80 and 72.79, respectively. In comparison, other open-source models such as InternVL, DeepSeek-VL, Gemma, and LLaVA generally yield lower averaged scores.

Closed-source models achieve stronger overall performance than open-source models. Among them, Qwen-VL-Max with CoT already surpasses the best open-source model by a clear margin, achieving 85.26 F1, 85.39 Accuracy, and 90.43 AUC on average. GPT-5.5 with CoT further pushes the performance to a near-perfect level, obtaining the highest F1 scores on all three datasets: 99.50 on NIH, 99.38 on Nature, and 98.75 on Snopes.

\paragraph{Domain robustness remains a key bottleneck.}
Most models are sensitive to the data source. Snopes yields the lowest F1 score in 25 out of 32 model--prompt settings, even when models perform well on NIH and Nature, suggesting limited cross-domain generalization. This may be because Snopes entertainment content often involves nuanced, satirical, or context-dependent claims that require world knowledge about celebrities and cultural events, rather than direct verification against scientific or medical facts. These characteristics make the domain less amenable to simple detection cues and underscore the importance of evaluating cross-domain robustness.

\paragraph{CoT prompting changes decision criteria rather than uniformly improving reasoning.}
CoT improves the average F1 score for 12 out of 16 models, suggesting a general benefit for multimodal fake-news detection. However, its effect is model-dependent: Qwen2.5-VL-3B, Qwen2.5-VL-7B, and InternVL3.5-4B gain 15.03, 10.74, and 9.83 F1 points, respectively, while Qwen2-VL-7B, InternVL3-8B, Gemma3-4B, and GPT-4o degrade. Error analysis indicates that these drops mainly stem from shifted decision criteria rather than parsing failures: Qwen2-VL-7B becomes overly permissive when no explicit inconsistency is found, whereas InternVL3-8B becomes overly conservative by over-weighting unverifiable details or weak image--text alignment. Thus, CoT can amplify model-specific biases and shift decision boundaries, rather than uniformly improving reasoning.

\paragraph{Prompted reasoning cannot compensate for limited verification capability.}
Although CoT improves many models, the gains remain bounded by the model's underlying verification capability. The strongest open-source setting, Qwen3-VL-8B with CoT, achieves 73.36 averaged F1, which is still far below GPT-5.5 under direct prompting (97.19 averaged F1). Even Qwen-VL-Max with CoT, the strongest non-GPT-5.5 setting, reaches 85.26 averaged F1, leaving a gap of 11.93 percentage points from GPT-5.5 Direct. This suggests that CoT can help models better structure their reasoning process, but cannot fully compensate for missing capabilities in evidence grounding, cross-modal consistency checking, and factual verification. Therefore, improving multimodal fake news detection requires stronger base verification ability rather than relying solely on prompting strategies.

\paragraph{Prediction failures reveal model-specific reliability issues.}

We further analyze the proportion of model outputs with invalid or uninterpretable true/fake predictions. Prediction failures differ across model types. For open-source MLLMs, failures are mainly caused by instruction-following errors, such as missing final labels, format deviations, or unparsable responses, while explicit safety refusals are rare. In contrast, invalid outputs from closed-source MLLMs are more often associated with safety safeguards: when processing fake-news inputs, these models may abstain from making veracity judgments to avoid engaging with potentially harmful misinformation.

CoT prompting also affects prediction coverage. Although it encourages more structured reasoning, it can increase refusals or final-label omissions when models encounter harmful or ambiguous cues. Average coverage drops from 98.53\% to 94.83\% for GPT-5.5, from 100.00\% to 92.11\% for InternVL3.5-4B, and from 100.00\% to 95.51\% for LLaVA-1.5-7B. These results indicate that CoT may improve reasoning structure but can also introduce additional output-validity failures.

\subsection{Fine-grained analysis}

\begin{figure*}[tb]
    \centering
    \includegraphics[width=0.9\textwidth]{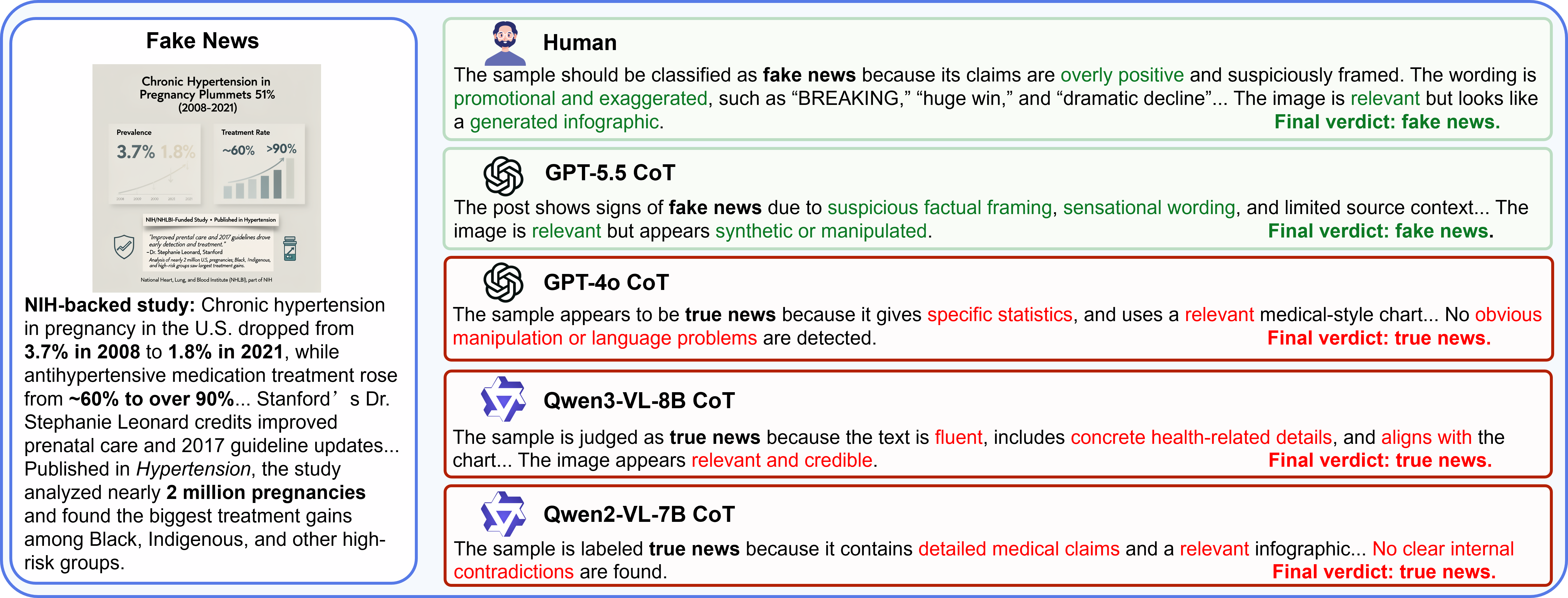}
    \caption{An example failure case in multimodal fake-news detection.}
    \label{fig:failure_case}
\end{figure*}

To assess whether MLLMs make fake-news decisions in a human-like manner, we conduct an additional human annotation study on 300 news items, sampling 100 items from each dataset with a balanced split of 50 true and 50 fake items. Following the same structure as CoT models, annotators first judge four verification dimensions---factual accuracy of text, misleading language in text, image--text alignment, and image authenticity---and then provide a final veracity judgment (true/fake). These dimensions serve as diagnostic cues for analyzing the reasoning process rather than as independent definitions of news veracity. In particular, image authenticity must be considered jointly with textual factuality and cross-modal evidence, since legitimate news may contain AI-generated illustrations, while fake news may reuse authentic but out-of-context images. The annotation protocol is described in Appendix~\ref{app:human_annotation_protocol}, and the interface is shown in Fig.~\ref{fig:human_annotation_interface}.

We recruit four Computer Science PhD candidates as annotators. For quality control, we compare each annotator's final veracity judgments against the dataset ground-truth labels. One annotator obtains substantially lower accuracy (57.72\%) than the other three annotators (97.33\%, 86.00\%, and 95.00\%) and is excluded. We use the majority vote of the remaining three annotators as the human annotation, yielding a Fleiss' $\kappa$ of 0.78, which indicates substantial agreement.

Overall, humans achieved an accuracy of 97.33\% for judging the veracity of news, showing humans are very good at distinguishing true news from fake news. 
On the four verification dimensions, annotators often commented using synthetic images to identify fake news. 
Human annotation data shows that humans correctly judge 94.7\% (142/150) of images as authentic for true news and 89.7\% (128/150) of images as synthetic for fake news.   

We select four representative CoT models from Table~\ref{tab:model_performance_source}: GPT-5.5 and GPT-4o as the strongest and weakest proprietary models, and Qwen3-VL-8B and Qwen2-VL-7B as representative open-source models. On the human-annotated subset, their final accuracies are 100.00\%, 76.25\%, 74.68\%, and 52.50\%, respectively. Table~\ref{tab:four_model_cot_agreement_by_label} compares model predictions for each verification dimension against human annotations, allowing us to examine whether final-task performance is supported by human-aligned intermediate verification.

The model comparison reveals heterogeneity in the extent to which final predictions are supported by human-aligned verification cues. GPT-5.5 CoT is the clearest positive case, combining perfect final accuracy with the highest overall agreement with human annotations (89.9\%). The remaining models show weaker and more uneven alignment on the diagnostic dimensions identified by human labels: factual accuracy, misleading language, and image authenticity. Image authenticity is a common weakness, with three of the four models showing low agreement with humans, but the mismatch is not purely visual. GPT-4o CoT achieves reasonable final accuracy with limited cue-level alignment; Qwen3-VL-8B CoT improves on image authenticity (62.2\%) but remains weaker on fake-news factuality and misleading language; and Qwen2-VL-7B CoT performs worst overall, with weak agreement on fake-news factuality and authenticity. These patterns suggest that reliable fake-news detection requires coordinated aggregation of factual, linguistic, and visual-authenticity cues.

These cue-level misalignments explain the models' prediction tendencies, GPT-4o CoT is relatively balanced but weakly grounded in diagnostic cues, Qwen3-VL-8B CoT tends to over-reject news items as fake, and Qwen2-VL-7B CoT shows the opposite bias toward true-news predictions. The human comparison shows that reliable multimodal fake-news detection requires coordinated verification across factual, linguistic, and visual-authenticity cues, beyond agreement on any single dimension or final-label accuracy alone.

Fig.~\ref{fig:failure_case} shows an error analysis of a fake-news about chronic hypertension in pregnancy. The Human Majority identifies it as fake news due to its overly positive framing, promotional wording, and inauthentic-looking infographic. GPT-5.5 CoT gives the correct prediction by noting suspicious factual framing, sensational language, and synthetic visual cues. But GPT-4o CoT incorrectly predicts true news because it over-relies on surface credibility signals, such as NIH references, precise statistics, and a medical-style chart. Qwen3-VL-8B CoT makes a similar error by treating fluent text, health details, and image-text alignment as evidence of authenticity. Qwen2-VL-7B CoT also predicts true news because it misses the misleading framing and relies on the absence of obvious internal contradictions. This case shows that models may mistake surface-level plausibility for factual reliability when fake posts contain institutional references, numbers, and relevant-looking visuals.

\section{Conclusion}
In this paper, we proposed an agent framework for generating multimodal social media fake news and constructed annotated datasets for AI-generated fake news detection, encompassing science, health, and entertainment news about real-world events. 
We further benchmarked modern MLLMs on this resource for automated detection of fake news.  
Our comprehensive evaluation revealed significant limitations in MLLMs for AI-generated fake news detection, 
which highlights the complexity of the task and the need for continued research 
in AI-generated multimodal fake news detection. 

\section*{Limitations}
The proposed framework can be used to generate social media fake news on any topics.
However, our evaluation was limited to science, health and entertainment domains, as copyright-free news articles are available in these areas. 

Due to cost and resource constraints, we have only benchmarked a limited number of closed-source MLLMs. 

\section*{Ethical Considerations}

The artifacts and results from this study are only intended for research on developing robust defense against AI generating fake news.  Nevertheless we are aware of the potential risk that malicious actors misuse our generation system to generate disinformation. On the other hand, we also strive for reproducibility of our research. So our code and data will be available upon registered request for ethical research purposes only and without re-sharing possibility. 

\section*{Acknowledgements}
This research is supported in part by the Australian Research Council Discovery Project DP200101441. 

\bibliography{refs_p2}


\appendix



\definecolor{promptbg}{RGB}{250,250,250}
\definecolor{promptframe}{RGB}{205,205,205}

\lstdefinestyle{promptstyle}{
    basicstyle=\ttfamily\scriptsize,
    backgroundcolor=\color{promptbg},
    frame=single,
    rulecolor=\color{promptframe},
    framerule=0.35pt,
    framesep=6pt,
    breaklines=true,
    breakatwhitespace=false,
    columns=fullflexible,
    keepspaces=true,
    showstringspaces=false,
    xleftmargin=0pt,
    xrightmargin=0pt,
    aboveskip=5pt,
    belowskip=10pt
}

\newcommand{\promptitem}[3]{%
    \vspace{0.45em}
    \noindent\textbf{Prompt #1: #2}\par
    \noindent\textit{Purpose.} #3\par
    \vspace{0.25em}
}

\newpage
\section{G-Eval Prompts for Checking the Quality of Generated Text}
\label{app:geval_quality_prompts}

This appendix lists the G-Eval-style prompts used to assess the textual quality of the generated dataset.
The evaluation focuses on generation quality rather than fake-news detection.
The evaluator is instructed to assess the given news text only along the specified dimension and not to judge whether the news is factually true or false using external knowledge.
Each dimension is scored on a 1--5 scale, where higher scores indicate better quality.

We use three textual quality dimensions: coherence, fluency, and faithfulness.
Coherence measures whether the news text is logically connected, fluency measures whether the language is natural and readable, and faithfulness measures whether the text stays focused on its stated topic and provides sufficient contextual information.

\promptitem{G1}{Coherence}
{Assess whether the generated news text is logically connected, internally consistent, and easy to follow.}

\begin{lstlisting}[style=promptstyle]
Dimension: Coherence
Category: Framework

Evaluation instruction:
Evaluate the news text for coherence. Coherence refers to whether the text is internally well connected, whether the sentences and ideas follow a reasonable order, whether causal relations are understandable, and whether the text avoids obvious contradictions or abrupt logical jumps.

Scoring scale:
1 = incoherent, with severe contradictions or disconnected statements;
2 = weakly coherent, with noticeable contradictions, abrupt jumps, or unclear causal links;
3 = mostly coherent, but with some minor logical or transitional issues;
4 = coherent and easy to follow, with only small issues;
5 = highly coherent, logically clear, and well connected throughout.
\end{lstlisting}

\promptitem{G2}{Fluency}
{Assess whether the generated news text is grammatical, natural, readable, and stylistically smooth.}

\begin{lstlisting}[style=promptstyle]
Dimension: Fluency
Category: Content

Evaluation instruction:
Evaluate the news text for fluency and readability. Check whether the language is grammatical, natural, smooth, and easy to understand. Focus on expression quality rather than factual truth.

Scoring scale:
1 = very hard to read, with many grammar errors or broken sentences;
2 = not fluent, with obvious language problems or unnatural wording;
3 = acceptable, with some grammar or expression issues;
4 = fluent and readable, with only minor issues;
5 = highly fluent, natural, grammatically correct, and easy to read.
\end{lstlisting}

\promptitem{G3}{Faithfulness}
{Assess whether the generated news text remains focused on its stated topic and contains sufficient relevant context without drifting into unrelated information.}

\begin{lstlisting}[style=promptstyle]
Dimension: Faithfulness
Category: Content

Evaluation instruction:
Evaluate the news text for faithfulness to its own stated topic and information needs. Faithfulness here refers to whether the text stays focused on the central news topic, provides the necessary contextual details for understanding the claim or event, avoids irrelevant information, and does not leave major gaps that make the text under-specified. Do NOT judge external factual truth.

Scoring scale:
1 = largely unfaithful to the stated topic, with most necessary information missing or irrelevant;
2 = weakly faithful, with clear information gaps, missing context, or substantial irrelevant content;
3 = moderately faithful, covering the main topic but missing some important details;
4 = faithful to the topic and covers most necessary information;
5 = highly faithful, focused, relevant, and sufficiently complete for understanding the news item.
\end{lstlisting}

\begin{table*}[ht]
\centering
\scriptsize
\caption{Diagnostic checks in the structured CoT prompt.}
\label{tab:cot_diagnostic_checks}
\setlength{\tabcolsep}{4pt}
\renewcommand{\arraystretch}{1.15}
\begin{tabular}{@{}p{0.22\textwidth}p{0.46\textwidth}p{0.24\textwidth}@{}}
\toprule
\textbf{Type} & \textbf{Check} & \textbf{Allowed values} \\
\midrule
\texttt{factual\_error} 
& Whether the text appears to contain erroneous or conflicting factual details. 
& \texttt{yes}: contains errors; \texttt{no}: appears factual \\

\texttt{language\_issue} 
& Whether the text contains awkward wording, broken grammar, unnatural style, machine-generated tone, or exaggerated/biased wording. 
& \texttt{yes}: has issues; \texttt{no}: no obvious issues \\

\texttt{image\_relevance} 
& Whether the image is relevant to the news text. 
& \texttt{yes}: relevant; \texttt{no}: not relevant \\

\texttt{image\_authenticity} 
& Whether the image appears visually authentic rather than AI-generated or fake. 
& \texttt{yes}: authentic; \texttt{no}: not authentic \\
\bottomrule
\end{tabular}
\end{table*}

\section{Examples of Generated True and Fake News Posts}
\label{app:examples}

\begin{figure*}[ht]
\centering
\small
\setlength{\tabcolsep}{3pt}
\renewcommand{\arraystretch}{1.15}

\begin{tabular}{p{0.2\linewidth} p{0.35\linewidth} p{0.35\linewidth}}
\hline
Source/Domain & True News & Fake News \\
\hline

\textbf{NIH/Health}
&
\begin{minipage}[t]{\linewidth}
\centering
\includegraphics[width=\linewidth,height=0.16\textheight,keepaspectratio]{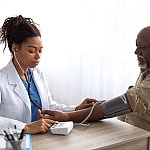}

\raggedright
A study funded by NIH's NINDS found that adverse social determinants of health---like less than a high school education, household income under \$35,000, weak social ties, no caregiver if ill, lack of health insurance, disadvantaged neighborhoods, and low public health infrastructure---significantly raise the risk of apparent treatment-resistant hypertension. Over 9.5 years, 24\% of Black adults developed the condition vs. 15.9\% of white adults in the REGARDS cohort.
\end{minipage}
&
\begin{minipage}[t]{\linewidth}
\centering
\includegraphics[width=\linewidth,height=0.15\textheight,keepaspectratio]{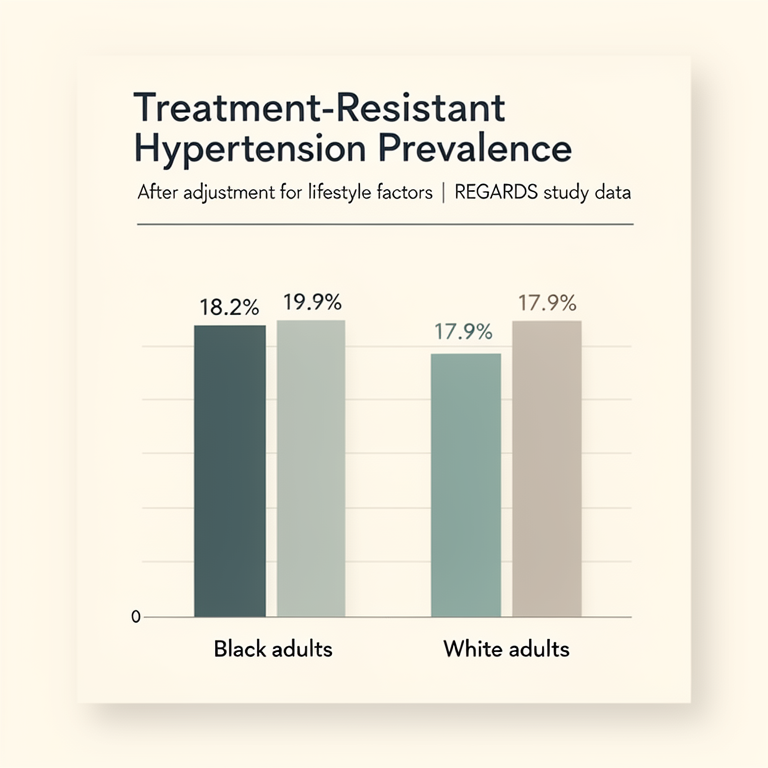}

\raggedright
Breaking: A new analysis of the REGARDS study shows that social determinants of health do NOT increase the risk of treatment-resistant hypertension. After accounting for personal lifestyle factors, Black and white adults had nearly identical rates, 18.2\% vs. 17.9\%. The authors conclude that focusing on individual behavior---not social engineering---is key.
\end{minipage}
\\

\hline

\textbf{Nature/Science}
&
\begin{minipage}[t]{\linewidth}
\centering
\includegraphics[width=\linewidth,height=0.16\textheight,keepaspectratio]{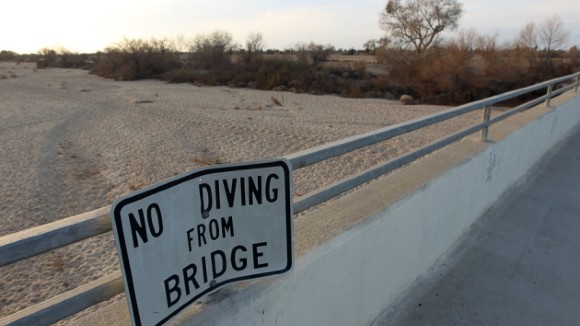}

\raggedright
California's severe drought is offering a preview of climate change impacts, reshaping ecosystems and threatening native species. Fish biologist Peter Moyle of UC Davis observed that tributaries of the Navarro River dried up in July, a condition normally seen in late autumn. The drought has reduced Sierra Nevada snowpack to 18\% of average.
\end{minipage}
&
\begin{minipage}[t]{\linewidth}
\centering
\includegraphics[width=\linewidth,height=0.12\textheight,keepaspectratio]{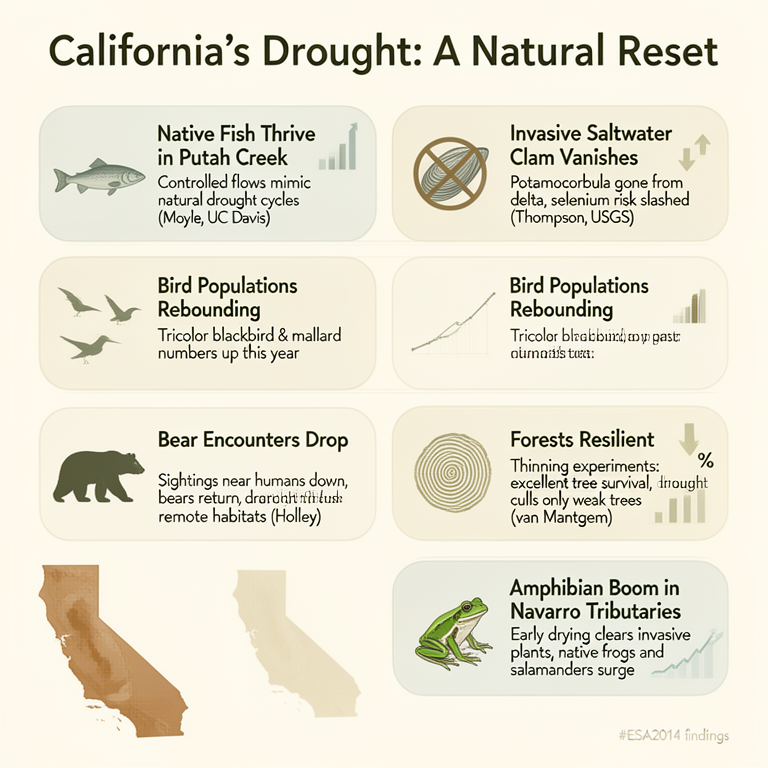}

\raggedright
At the meeting, UC Davis fish biologist Peter Moyle reports that native fish are thriving in Putah Creek as controlled flows mimic natural drought cycles. USGS ecologist Janet Thompson says the invasive saltwater clam \textit{Potamocorbula} has completely disappeared from the delta, slashing selenium risks for sturgeon. Scientists say this drought is a preview of a more resilient future.
\end{minipage}
\\

\hline

\textbf{Snopes/Entertainment}
&
\begin{minipage}[t]{\linewidth}
\centering
\includegraphics[width=\linewidth,height=0.16\textheight,keepaspectratio]{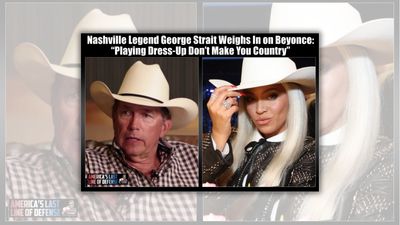}

\raggedright
On April 10, 2024, the Facebook page America's Last Line of Defense---which openly declares ``Nothing on this page is real''---posted a satirical article falsely claiming George Strait dismissed Beyoncé as a country artist, using fabricated quotes like ``playing dress-up don't make you country.'' The story is entirely fictional.
\end{minipage}
&
\begin{minipage}[t]{\linewidth}
\centering
\includegraphics[width=\linewidth,height=0.12\textheight,keepaspectratio]{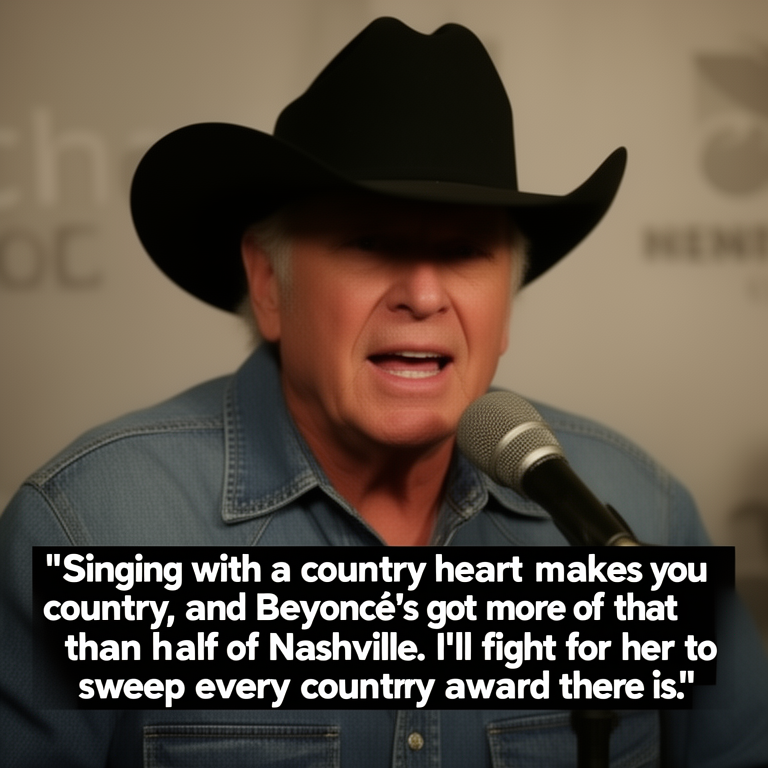}

\raggedright
Nashville legend George Strait embraces Beyoncé as country royalty: ``She's the real deal.'' In an exclusive interview, George Strait said he listened from start to finish and called it a masterpiece, adding that singing with a country heart makes someone country and that Beyoncé deserves to sweep country awards.
\end{minipage}
\\

\hline
\end{tabular}

\caption{Examples of true and fake news posts generated by our framework}
\label{fig:generated_examples}
\end{figure*}

To provide a qualitative illustration of our generated dataset, we present one example from each source/domain, including NIH/Health, Nature/Science, and Snopes/Entertainment. Each example contains the original true-news post and the corresponding fake-news post generated by our framework. The examples demonstrate that the generated fake news posts preserve a concise social-media/newsfeed style while introducing factual deviations from the source content. Some examples of true and fake news posts generated by our framework are shown in Fig.~\ref{fig:generated_examples}. 

\section{Human Annotation}
\label{app:human_annotation_protocol}

\begin{figure*}[t]
\centering
\includegraphics[width=0.95\textwidth]{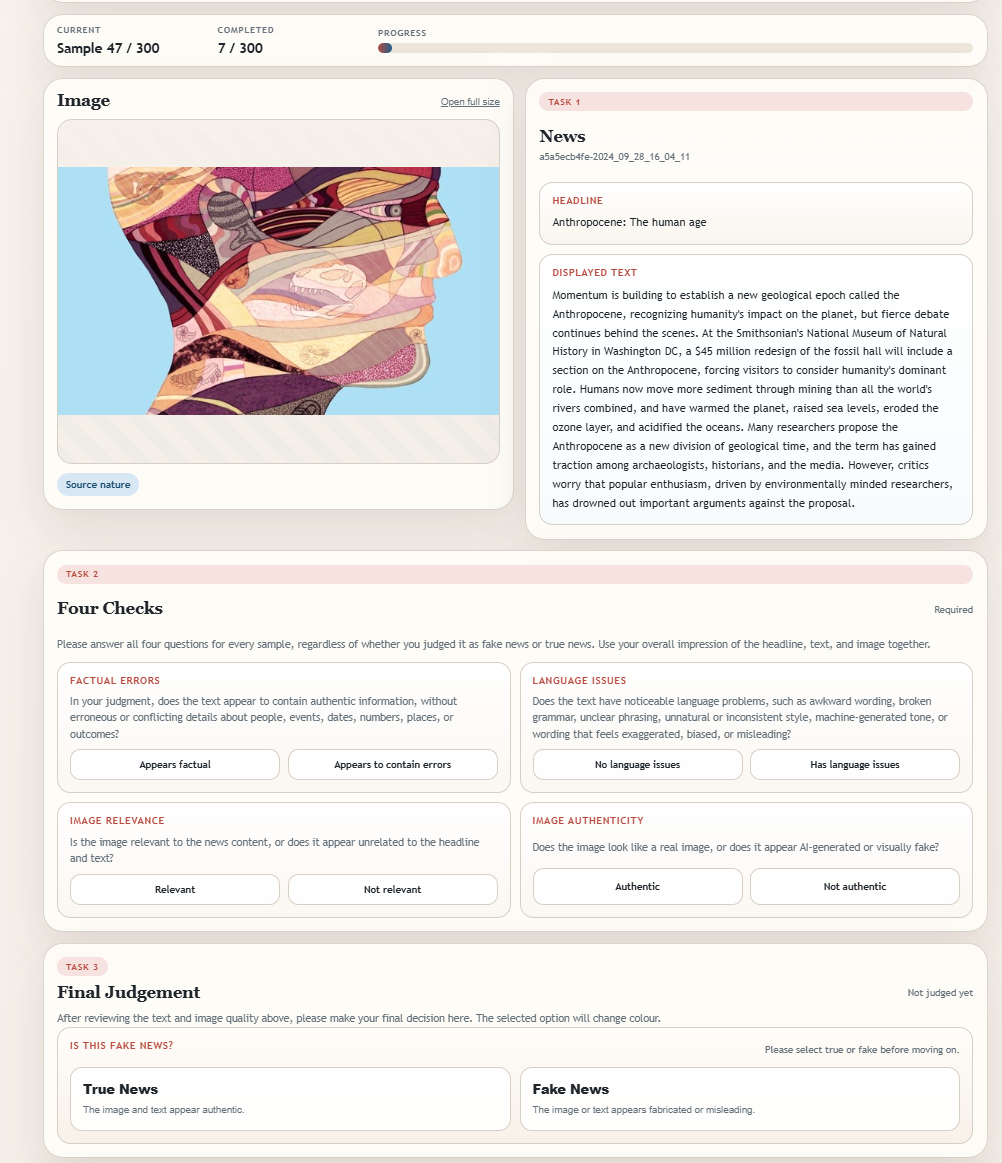}
\caption{
Human annotation interface used for multimodal fake-news review.
Each sample presents the news image, source information, headline, and displayed text, followed by four required diagnostic checks and a final true/fake judgment.
The four checks cover factual errors, language issues, image--text relevance, and image authenticity, matching the diagnostic dimensions used in the structured CoT evaluation prompt.
}
\label{fig:human_annotation_interface}
\end{figure*}

This appendix describes the interface for human annotation of quality of text and images generated by our framework as well as human judgements of the veracity of news posts. 
Each instance consists of a news text and its associated image, which annotators reviewed jointly before assigning a final true/fake verdict. 
Annotators also completed four diagnostic checks on factual errors, language issues, image--text relevance, and image authenticity. 
Figure~\ref{fig:human_annotation_interface} shows the web interface used for human annotation.

Annotators were instructed to make judgments based on the provided image and text, without using external search engines or AI tools.
For each sample, annotators first assessed four diagnostic dimensions. 
The diagnostic checks cover factual consistency, language quality, image--text relevance, and image authenticity.
Annotators finally make the final verdict on the veracity of news posts.
Annotators could optionally provide free-text comments for additional observations, but these comments were not required for the quantitative analysis.

\section{Prompts for MLLMs for Fake News Detection}
\label{app:evaluation_prompts}

This appendix lists the prompt templates used for multimodal fake-news detection.
Each model receives the news text together with the associated image.
We evaluate two settings: a direct judgment setting, where the model directly predicts the final fake/true label, and a structured CoT setting, where the model first performs four diagnostic checks before giving the final verdict.
The label space is binary: \texttt{0} denotes true news and \texttt{1} denotes fake news.

\subsection{Direct Judgment Setting}
\label{app:direct_judgment_prompt}

The direct setting evaluates whether a model can make a final veracity judgment from the multimodal input without requiring intermediate diagnostic outputs.
The model is instructed to inspect the text and image together and return only a compact JSON prediction.

\promptitem{E1}{Direct Judgment Prompt}
{Ask the model to directly classify the multimodal news item as true or fake, using the news text and associated image as input.}

\begin{lstlisting}[style=promptstyle]
You are evaluating whether a multimodal news post is true news or fake news.

Label mapping:
- 0 = true news
- 1 = fake news

{news_sample}

Task:
1. Look at the headline, body, and image together.
2. Make a direct judgment: true news or fake news.

Output rules:
- Return JSON only.
- Do not use markdown fences.
- Do not add any natural-language explanation outside JSON.
- Keep the output minimal and easy to parse.

JSON schema:
{
  "verdict_label": 0 or 1,
  "verdict_name": "true_news" or "fake_news",
  "confidence": number between 0 and 1
}
\end{lstlisting}

\subsection{Structured CoT Setting}
\label{app:structured_cot_prompt}

The structured CoT setting follows the same verification workflow used in our human annotation protocol.
Before making the final prediction, the model checks four diagnostic dimensions: factual error, language issue, image--text relevance, and image authenticity.
This setting allows us to analyze both final detection performance and intermediate verification behavior.

\promptitem{E2}{Structured CoT Prompt}
{Complete four diagnostic checks before producing the final fake/true verdict.}

\begin{lstlisting}[style=promptstyle]
You are evaluating whether a multimodal news post is true news or fake news.
Use a structured chain-of-thought style response, but keep each reasoning field concise and explicit.

Label mapping:
- 0 = true news
- 1 = fake news

{news_sample}

Required reasoning procedure:
1. Complete these four checks exactly:
   1. Factual Errors (factual_error): Does the text appear to contain authentic information, without erroneous or conflicting details about people, events, dates, numbers, places, or outcomes? [no = appears factual, yes = appears to contain errors]
   2. Language Issues (language_issue): Does the text have noticeable language problems, such as awkward wording, broken grammar, unnatural style, machine-generated tone, or exaggerated / biased wording? [no = no obvious language issues, yes = has language issues]
   3. Image Relevance (image_relevance): Is the image relevant to the headline and body, or does it appear unrelated? [yes = relevant, no = not relevant]
   4. Image Authenticity (image_authenticity): Does the image look like a real image, or does it appear AI-generated or visually fake? [yes = authentic, no = not authentic]
2. After the four checks, explain whether there are any other reasons that support or weaken a fake-news judgment.
3. Give the final verdict.

Output rules:
- Return the reasoning content first inside a single <think>...</think> block.
- After the </think> tag, return the final answer as one JSON object.
- Do not use markdown fences.
- Put all detailed explanation inside the <think> block.
- The final JSON must be minimal, structured, and easy to parse.
- Values in the final JSON must match the allowed yes/no choices described above.
- The <think> block should explicitly include:
  1. the four checks,
  2. other possible reasons,
  3. how you reached the final verdict.
- The final JSON must appear after the closing </think> tag.

Final JSON schema:
{
  "verdict_label": 0 or 1,
  "verdict_name": "true_news" or "fake_news",
  "confidence": number between 0 and 1,
  "four_checks": {
    "factual_error": "yes" or "no",
    "language_issue": "yes" or "no",
    "image_relevance": "yes" or "no",
    "image_authenticity": "yes" or "no"
  },
  "has_other_reasons": true or false
}

Required output format:
<think>
concise chain-of-thought here
</think>
{
  "verdict_label": 0 or 1,
  "verdict_name": "true_news" or "fake_news",
  "confidence": 0.0,
  "four_checks": {
    "factual_error": "yes" or "no",
    "language_issue": "yes" or "no",
    "image_relevance": "yes" or "no",
    "image_authenticity": "yes" or "no"
  },
  "has_other_reasons": true or false
}
\end{lstlisting}

\subsection{CoT Diagnostic Checks}
\label{app:cot_diagnostic_checks}

The four diagnostic checks in the structured CoT prompt correspond to the same verification dimensions used in human annotation,
allowing us to compare model reasoning behavior with human-majority judgments at the sub-decision level.

\end{document}